\documentclass[conference]{IEEEtran}
\IEEEoverridecommandlockouts
\usepackage{cite}
\usepackage{amsmath,amssymb,amsfonts}
\usepackage{algorithmic}
\usepackage{graphicx}
\usepackage{textcomp}
\usepackage{xcolor}
\def\BibTeX{{\rm B\kern-.05em{\sc i\kern-.025em b}\kern-.08em
    T\kern-.1667em\lower.7ex\hbox{E}\kern-.125emX}}
\begin{document}

\title{A Multi-term and Multi-task Analyzing Framework for Affective Analysis in-the-wild}

\author{\IEEEauthorblockN{Sachihiro Youoku, Yuushi Toyoda, Takahisa Yamamoto, \\Junya Saito, Ryosuke Kawamura, Xiaoyu Mi, Kentaro Murase$^*$}
\IEEEauthorblockA{\textit{$^*$ Trusted AI Project, Artificial Intelligence Laboratory} \\
\textit{Fujitsu Laboratories Ltd., Kanagawa, Japan}\\}

}

\maketitle

\begin{abstract}
Human affective recognition is an important factor in human-computer interaction. However, the method development with in-the-wild data is not yet accurate enough for practical usage. In this paper, we introduce the affective recognition method focusing on valence-arousal (VA) and expression (EXP) that was submitted to the Affective Behavior Analysis in-the-wild (ABAW) 2020 Contest. Since we considered that affective behaviors have many observable features that have their own time frames, we introduced multiple optimized time windows (short-term, middle-term, and long-term) into our analyzing framework for extracting feature parameters from video data. Moreover, multiple modality data are used, including action units, head poses, gaze, posture, and ResNet 50 or Efficient NET features, and are optimized during the extraction of these features.
Then, we generated affective recognition models for each time window and ensembled these models together. Also, we fussed the valence, arousal, and expression models together to enable the multi-task learning, considering the fact that the basic psychological states behind facial expressions are closely related to each another. In the validation set, our model achieved a valence-arousal score of 0.498 and a facial expression score of 0.471. These verification results reveal that our proposed framework can improve estimation accuracy and robustness effectively.

\end{abstract}

\begin{IEEEkeywords}
Action Unit, Valence-Arousal, Emotional Expression, multi-term, multi-task
\end{IEEEkeywords}

\section{INTRODUCTION}

\begin{figure}[t]
\centerline{\includegraphics[width=8.5cm]{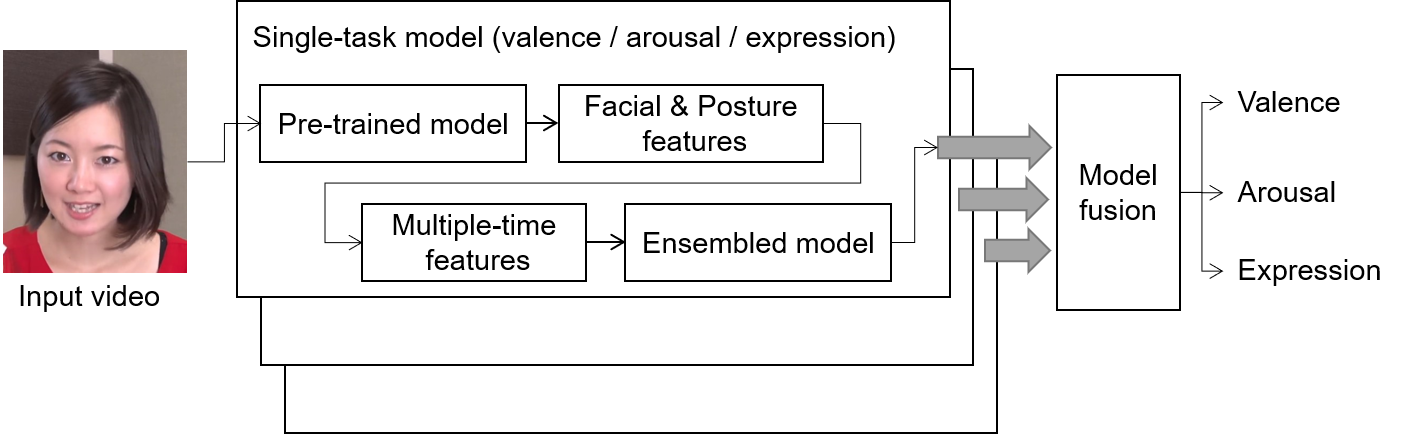}}
\caption{Overview of the proposed method for predicting Valence, Arousal, Expression.}
\label{fig:overview}
\end{figure}

Human affective recognition is an important factor in human-computer interaction. It is expected to contribute to a wide range of fields such as remote healthcare, learning, driver state monitoring, and so on. Many methods to express human mental state have been studied, of which “categorical emotion classification” and “Valence-Arousal” are the most commonly used methods. For the emotional category, the famous six basic emotional expressions\cite{c1}\cite{c2} proposed by Ekman and Friesen are popular. Ekman et al. classify emotions as "anger, disgust, fear, happiness, sadness, surprise". Another way to express emotions is the emotional circumplex model\cite{c3} developed by Russell. The circumflex model, human emotions are mapped in a two-dimensional plane using two orthogonal axes of the valence axis and arousal axis.
Recently, D. Kollias has provided a large scale in-the-wild dataset, Aff-Wild2\cite{c4}\cite{c5}. Aff-wild2 is an extended version of Aff-wild\cite{c6}. this dataset has used actual videos including a wide range of content (different age, ethnicity, lighting conditions, location, image quality, etc.) collected from YouTube. And multiple labels such as 7 emotion classifications (6 basic emotion expressions + Neutral), Valence-Arousal, Action-unit (based on Facial action coding system (FACS)\cite{c7} have been annotated to the video.
In this paper, we propose a fusion model that uses multiple time scale features and different recognition tasks. Figure \ref{fig:overview} shows the framework of the fusion model.
When the video data is received, facial features and posture features are extracted by using the pre-trained models. Multiple modality features including action unit, head-pose, gaze, posture, as well as features from ResNet50 or Efficient NET are extracted and selected optimally. These features are then calculated over short-term, middle-term, and long-term time window to convert them into multi-term features.
A model for a single recognition task of Valence, or Arousal, or Expression is constructed by ensembling the multiple models with different time windows constructed by using different term features respectively. Furthermore, the final predictive model is generated by fusing other recognition task models. This paper will describe the detail of those model construction and their verification results.

\section{RELATED WORK}

Estimating not only the occurrence of emotions but also the intensity of them is a concern that has been studied for many years. In recent years, Van Tong Huin et al. have proposed a method for estimating the regression of engagement, that is strongly related to emotions, with high accuracy by ensembling Action-unit features obtained from Openface\cite{c8} and image features obtained from ResNet50\cite{c9} in "the 6th Wild Challenge in Emotional Recognition (EmotiW 2019)"\cite{c10}. Similarly, Zhiguang Zhou et al. estimate the regression problem of engagement with high accuracy by ensembling Action-unit features obtained from Openface and posture features obtained from Openpose \cite{c11}\cite{c12}. Ensembling weak models are one of the effective ways to improve emotion estimation accuracy.
Also, Nigel Bosch, Sidney D’Mello et al. investigated the relationship between time windows and classification performance in emotion classification. Their result showed that some emotions need different time windows for model construction. For example, “De-lighted” estimation model using a short time window shows better performance, and “Confused” estimation motel a long time window for higher performance\cite{c13}. With regard to the impact of various recognition tasks, D. Kollias et al. have shown that it can improve the performance comparing to each single task model to combine the task models of action-unit detection, emotional classification, and valence-arousal estimation into a fused model\cite{c14}\cite{c15}.

\section{METHODOLOGY}

In this section, we introduce our proposed method that combines multiple timescales and multiple recognition tasks. The method consists of pre-processing, multi-term model, and model fusion of multi-task.

\subsection{Visual Data Pre-processing}

First, as shown in Fig. 2, the facial expression features and posture features are extracted from each video. There are two types of facial features, one obtained from Openface and the other obtained from ResNet50\cite{c9} or EfficientNET\cite{c20}. From Openface, 49 dimensions consisting of Action unit Intensity (17 dimensions), Action unit Occurrence (18 dimensions), Head-pose (6 dimensions), Gaze features (8 dimensions) are acquired as the features F1. From ResNet50, after acquiring the 2048-dimensional image features F2, the features which have been dimensionally reduced to 200 dimensions by principal component analysis (PCA) are obtained as F2’. From EfficientNet, after acquiring the 2048-dimensional image features F2, the features which have been dimensionally reduced to 300 dimensions by principal component analysis (PCA) are obtained as F2’. The posture features are obtained from Openpose. For Openpose, 75-dimensional skeleton features (25-dimensional x 3-axis) noted as F3 are used.

Next, the features using multiple time windows for the features F1, F2’and F3 are computed, since we think that feature changes in emotions and facial expressions should be calculated on different time windows. For example, the yawning mouth feature is extracted in a long window, and the feature that raises the eyebrows with surprise is extracted in a short window. There are multiple types of time frames: short-term, middle-term, and long-term, which are used to extract various features. Furthermore, the features of each time window (Fs, Fm, Fl) consist of the following components.

\begin{itemize}
  \item Average value
  \item Standard deviation
  \item Maximum change width (maximum value - minimum value)
  \item Slope (using least squares method)
\end{itemize}

Similarly, for each time window, a label (Ls, Lm, Ll) is newly generated using the given annotations. The method of generating the label depends on the task and is as follows. 

\begin{itemize}
  \item Valence, Aroual: Average value of the given annotations
  \item Expression: Mode of the given annotations
\end{itemize}

\begin{figure}[t]
\centerline{\includegraphics[width=8.5cm]{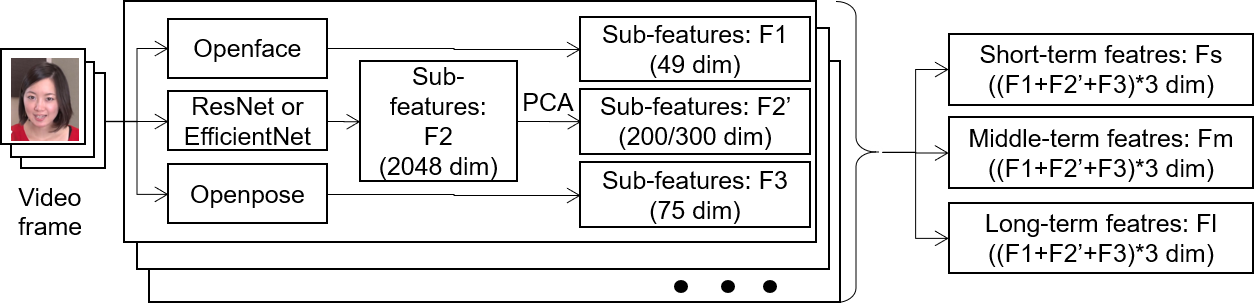}}
\caption{Pre-processing: feature engineering of multi model and multi time-scale data }
\label{fig:preproc}
\end{figure}

\subsection{Data Balancing}

\begin{figure}[t] 
\centerline{\includegraphics[width=4.0cm]{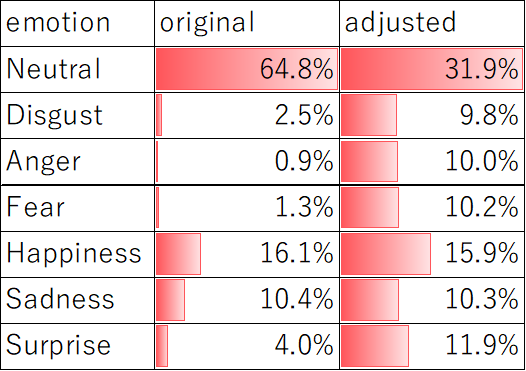}}
\caption{Expression data distributions }
\label{fig:balanceexp}
\end{figure}

\begin{figure}[t] 
\centerline{\includegraphics[width=8.5cm]{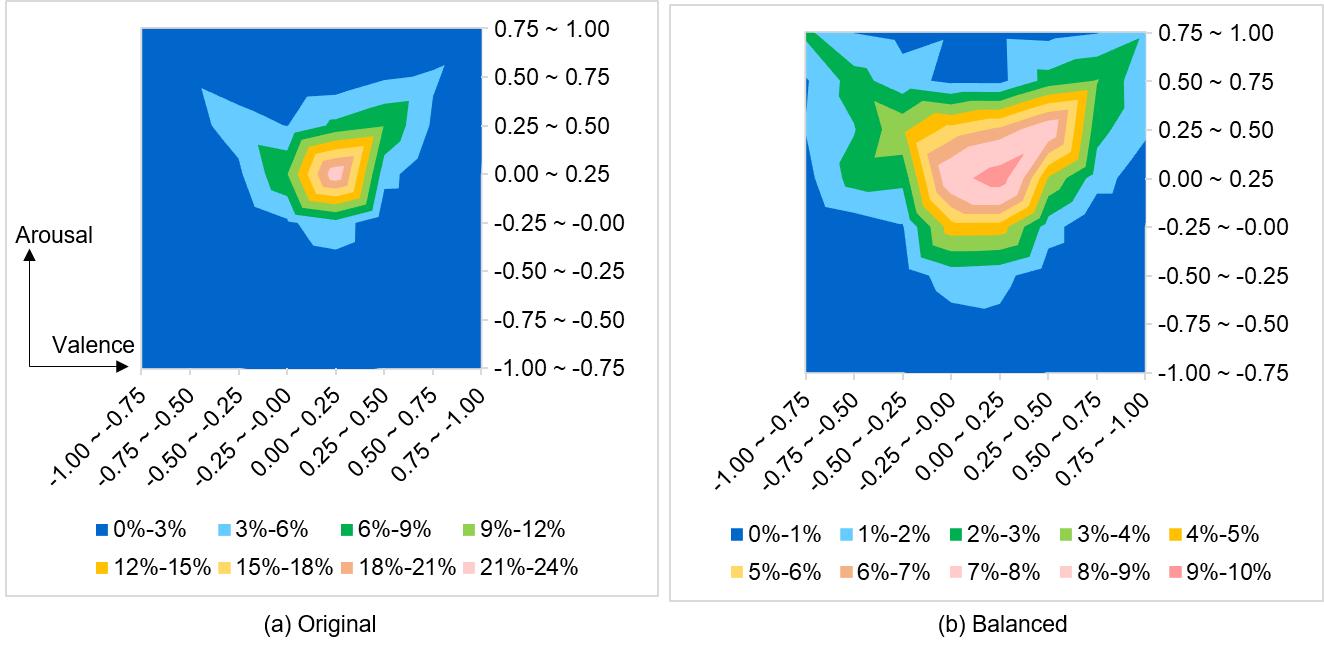}}
\caption{Valence-Arousal data distributions }
\label{fig:balanceva}
\end{figure}

It is important to address the data imbalance problem. 
In the annotations of the Aff-wild2 dataset, neutral accounts for more than 60\%, and anger and fear account for only about 1\%. In Valens-Arousal, more than 23\% of data is collected in the range of Valence: 0 to 0.25 and Arousal: 0 to 0.25. 
Therefore, the dataset needs to be balanced. Figure \ref{fig:balanceexp} shows the distribution of expression data after the balancing, and Figure \ref{fig:balanceva} shows the distribution of the Valens-Arousal data after the balancing. For facial expression data, the data is balanced by halving the number of neutral data and duplicating other emotional data. For Valens-Arousal data, the original data distribution is completely divided into 64 regions consisting of 8 Valence x 8 Arousal regions, then is balanced by halving the data in the center and duplicating the data elsewhere.

\subsection{Multi-term model}

\begin{figure}[t] 
\centerline{\includegraphics[width=8.5cm]{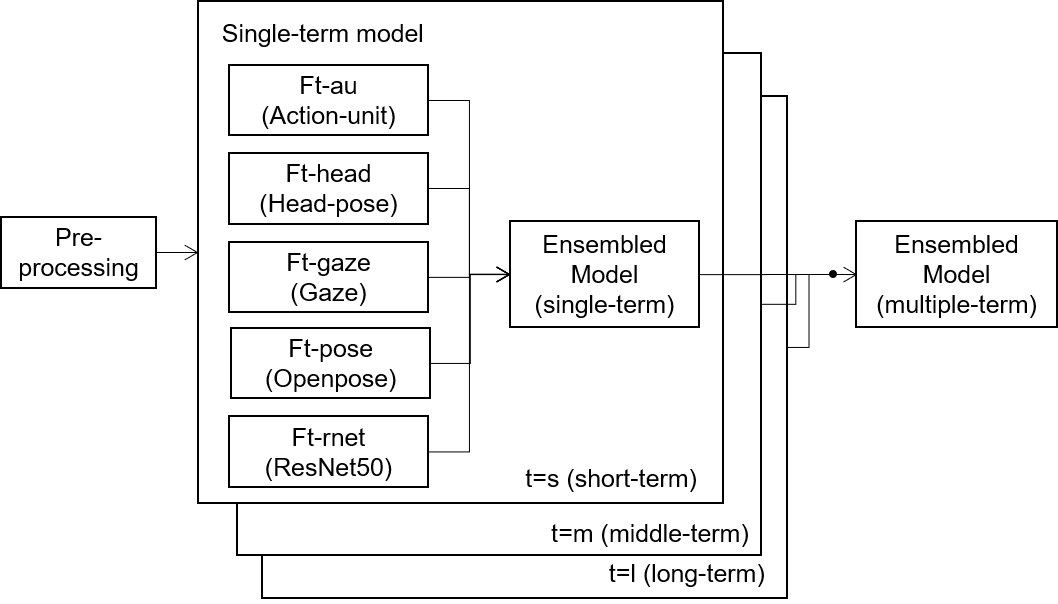}}
\caption{Multi-Term Model: ensembled short-term, middle-term, long-term model }
\label{fig:multiterm}
\end{figure}

The structure of the multi-term model is shown in Fig. \ref{fig:multiterm}. First, a single-term model is generated using Ft and Lt, which are the time window features and label described in the previous section (t is the target time window). In the single-term model, the feature: Ft consists of such sub-features as AU feature: Ft-au, Head-pose feature: Ft-head, Gaze feature: Ft-gaze, Openpose feature: Ft-pose, ResNet50 feature: Ft-rnet or EfficientNet feature: Ft-enet. These sub-features have their own selected feature components respectively. And the sub-estimation models are generated individually according to each sub-feature using their different sub-feature components. The final performance of a single-term model is improved by ensembling those sub-estimation models together, especially for the estimation of emotion\cite{c10}\cite{c12}\cite{c18}\cite{c19}. Labels for Valence and Arousal use the values averaged in each time window. In other words, the single term models estimate the trend of short-term, middle-term, long-term Valence and Arousal respectively.
Then, the model of single-task noted as Msingle-task recognizing the Expression, Valence and Arousal state individually, is generated by ensembling the three single-term models (Ms, Mm, Ml) according to short-term, middle-term, and long-term respectively. All the label values use the short-term window label values. This is because the short-term window label values are comparable to frame label values.

\subsection{Model fusion of multi-task}

\begin{figure}[t] 
\centerline{\includegraphics[width=8.5cm]{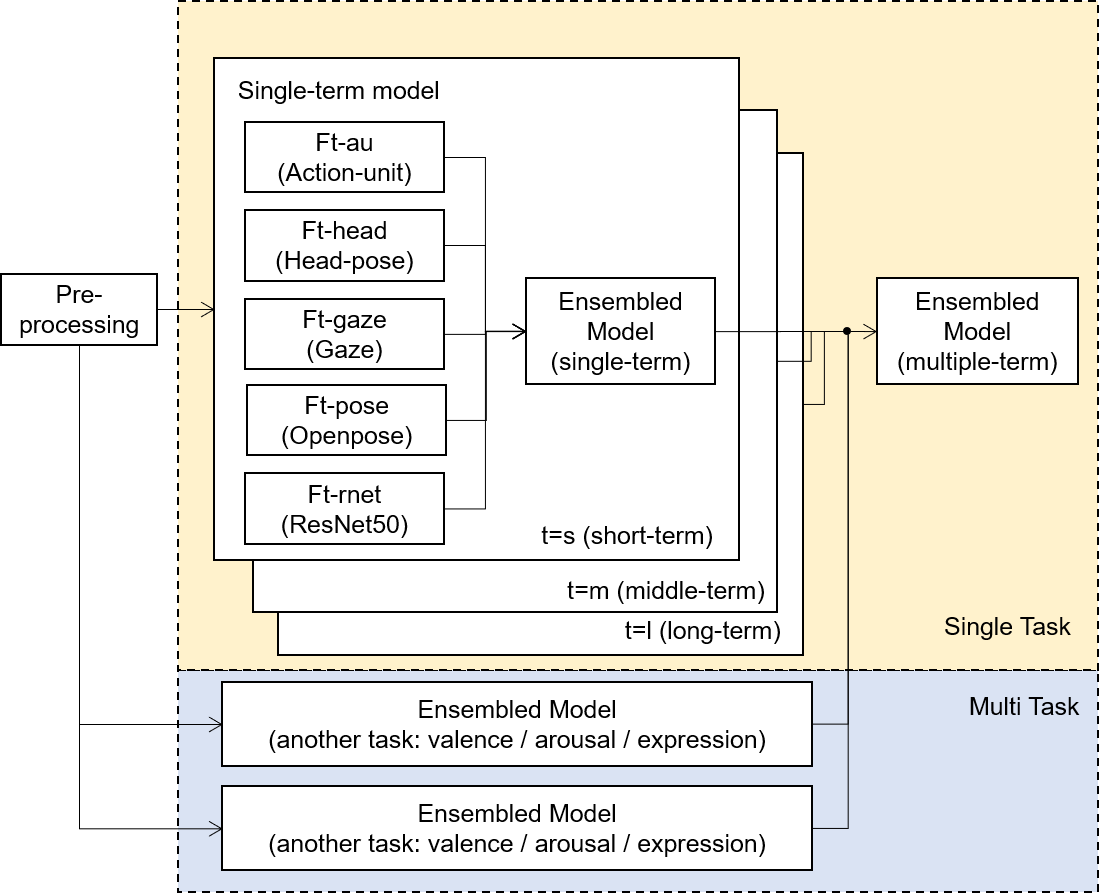}}
\caption{Multi Task Model: fusion multi-term model for taregt task and models for other tasks }
\label{fig:multitask}
\end{figure}

It has been reported that the estimation performance of the target task is improved by using different task features\cite{c14}\cite{c15}. Therefore, as shown in Fig. \ref{fig:multitask}, a Fusion model is generated by incorporating the estimated values for other recognition tasks as features into the Multi-term model. Fusion model uses Multi-term models generated in each task (single task of valence, single task of arousal, single task of expression). The estimated value of the target task is generated by combining the estimated values of the three single-term models for the target task and the estimated values of the multi-term models for non-target tasks. For example, when Valence is targeted, the estimated values of the short-term, middle-term, and long-term models that estimate Valence, the estimated values of the Multi-term model that estimate Arousal, and the Multi-term that estimates Expression are combined to generate the final Valence estimation model.

\section{EXPERIMENTS}

\subsection{Implementation and Setup}

[Dataset]

We used the Aff-wild2 dataset\cite{c4}\cite{c5}. This contains 548 videos, and multiple annotations (Valence-Arousal, Expression, etc.) are added in frame units. This is currently the largest audio-visually annotated in-the-wild dataset. In this challenge, the following Training subjects, Validation subjects, and Test subjects data were provided from the data annotated with Valence-Arousal and Expression.
\begin{itemize}
  \item Valence-Arousal: 351, 71, 139 subjects in the training, validation, test
  \item Expression: 253, 70, 223 subjects in the training, validation, test
\end{itemize}
However, some videos may have multiple subjects in the frame. Since it is difficult to separate subjects from these videos, videos without these were used for training and validation. As a result, 341 training subjects and 65 validation subjects were used for Valence-Arousal task, and 244 training subjects and 68 validation subjects were used for Expression task. The test subjects were manually divided and used all.

[Evaluation Metric]

For Challenge-Track 1: Valence-Arousal estimation, ABAW Challenge used the Canonical Concordance Coefficient (CCC) metric as follows:
\begin{equation}
\label{eq:va}
 Score_{ccc. valence/arousal} = \frac{2s_{xy}}{s^2_x + s^2_y + (\bar{x} - \bar{y})^2} 
\end{equation}
where  $x$ and $y$ are the valence/arousal annotations and predicted values, $s_x$ and $s_y$ are their variances, $s_{xy}$ is the covariance, $\bar{x}$ and $\bar{y}$ are the mean values.
Total score of track 1 is Valence-Arousal the mean value of CCC in valence and arousal.
\begin{equation}
\label{eq:vatotal}
 Score_{ccc. total} = \frac{Score_{ccc. valence}+Score_{ccc. arousal}}{2} 
\end{equation}

For Challenge-Track 2: 7 Basic Expression Classification, ABAW Challenge used the accuracy and F1 score, and the score of track 2 is calculated as below equation:
\begin{equation}
\label{eq:exp}
 Score_{expression} = 0.67 * F_1 + 0.33 * Accuracy
\end{equation}

[Implementation]

Our framework was implemented by Jupyter lab. For feature extraction in pre-processing, Openface 2.2.0, ResNet50 or Ef-ficientNet, and Openpoe 1.5.1 were used, and standardization was performed for each feature. For short-term, middle-term, and long-term, we used time-windows of 1 second, 3 seconds, 6 seconds, and 12 seconds, respectively, and training data and validation data were generated with 0.2 second shift. As a result of preprocessing, 285,260 training data and 46,398 validation data were used for Valence-Arousal, and 285,260 training data and 46,398 validation data were used for Expression for each time window. We used LightGBM\cite{c16} to generate the learning model. In Valence-Arousal, \textit{regression} was used in \textit{objective} and the following custom functions was used in metric function. This is to balance CCC, which is the evaluation metric of this task, with MSE to minimize the error.

\begin{itemize}
  \item \textit{metric function} = 2 * CCC - MSE
  \begin{itemize}
  \item CCC: Canonical Concordance Coefficient
  \item MSE: Mean Squared Error
  \end{itemize}
\end{itemize}

In Expression, \textit{multiclass (7 classes)} was used in \textit{objective} and Eq. (\ref{eq:exp}), which is Evaluation Metric of this task, was used in \textit{metric function} as a custom function.
Among other parameters, \textit{num\_leaves}, \textit{learning\_rate}, \textit{max\_depth}, \textit{min\_child\_samples} were tuned by grid search.
The above tuning was performed for each model of AU, Head-pose, Gaze, Openpose, Resnet50 or EfficientNet, Single-term model, Multi-term model, and Multi-task model, and the final model was generated.
In addition, submissions up to 7 times are allowed in this challenge, so we generated models with the following patterns and validated it.
\begin{itemize}
  \item Using ResNet50 or EfficientNet
  \item Data balancing, or not
  \item Add 3-second window data as optional term to the 1-second, 6-second, and 12-second window data, or not
  \item Feature extraction (reduction of features to 50\% based on LoightGBM importances), or not
\end{itemize}

\subsection{Results and Discussion}

First, Table \ref{table:validation1} shows the comparison results on the validation set between models trained using only the features of single time window (short-term, middle-term, long-term), ensembled multiple time windows (multi-term), and fused different cognitive tasks (multi-task). 
The comparison results having the data balancing, 3-second window, and optimum feature extraction are also included in Table \ref{table:validation2}. The Expression Score is the result calculated based on Eq. (\ref{eq:exp}), and the Valence-Arousal Score is the result calculated based on Eq. (\ref{eq:va}) and Eq. (\ref{eq:vatotal}). As a result of the validation, it was confirmed that the score of Mult-term is higher than that of Single-term, and that the score of Multi-task is higher than that of Multi-term. 
In particular, the Score was significantly improved in the Multi-term model. We think that this is because the gestures expressing emotions have different time window features, such as yawning and raising eyebrows with surprise, and the model incorporates each feature effectively. 
Next, Table \ref{table:validation2} shows the validation results of various patterns of Multi-task. “Submit” in the table is the submission number for this challenge.

\begin{table}
\caption{SINGLE TERM / MULTIPLE TERM / MULTI-TASK COMPARISON RESULT ON THE VALIDATION SET}
\label{table:validation1}
\begin{center}
\begin{tabular}{clllp{0pt}clll}
\hline
\\[-6pt]
~ & \multicolumn{1}{c}{EXPR} & \multicolumn{1}{c}{~} & \multicolumn{3}{c}{Valence-Arousal}\\
\cline{2-2}\cline{4-6}\\[-6pt]
Method & \multicolumn{1}{c}{\textbf{Score}} & \multicolumn{1}{c}{~} 
& \multicolumn{1}{c}{Val.} & \multicolumn{1}{c}{Aro.}  & \multicolumn{1}{c}{\textbf{Score}}\\
\hline\hline
\\[-6pt]
Baseline \cite{c17} & 0.360 & ~ & 0.140 & 0.240 & 0.190 \\[2pt]
short-term & 0.364  & ~ & 0.327  & 0.417  & 0.402 \\[2pt]
Middle-term & 0.435  & ~ & 0.361  & 0.430  & 0.396 \\[2pt]
Long-term & 0.374  & ~ & 0.351  & 0.380  & 0.366 \\[2pt]
Multi-term & 0.426  & ~ & 0.455  & 0.504  & 0.480 \\[2pt]
Multi-task & \textbf{0.432}  & ~ & 0.455  & 0.508  & \textbf{0.482} \\[2pt]
\hline
\end{tabular}
\end{center}
\end{table}

\begin{table}
\caption{RESULTS ON THE VALIDATION SET OF ABAW CHALLENGE 2020}
\label{table:validation2}
\begin{center}
\begin{tabular}{clllp{0pt}clll}
\hline
\\[-6pt]
~ & \multicolumn{4}{c}{Pettern} & \multicolumn{1}{c}{EXPR} 
& \multicolumn{3}{c}{Valence-Arousal} \\ 
\cline{2-5}\cline{6-9}\\[-6pt]
Subjct & \multicolumn{1}{c}{enet.} & \multicolumn{1}{c}{bal.} 
& \multicolumn{1}{c}{ext.} & \multicolumn{1}{c}{3s.} 
& \multicolumn{1}{c}{\textbf{Score}} 
& \multicolumn{1}{c}{Val.} & \multicolumn{1}{c}{Aro.}  & \multicolumn{1}{c}{\textbf{Score}}\\
\hline\hline
\\[-6pt]
Base \cite{c17} &  &  &  &   & 0.360 & 0.140 & 0.240 & 0.190 \\[2pt]
\hline\\[-6pt]
submit. 1 &  &  &  &   & 0.432 & 0.455 & 0.508 & 0.482 \\[2pt]
\hline\\[-6pt]
submit. 2 & \checkmark &  &  &   & 0.435 & 0.429  & 0.502 & 0.466 \\[2pt]
\hline\\[-6pt]
submit. 3 &  &  & \checkmark & \checkmark & 0.462 & 0.500 & 0.489 & 0.495 \\[2pt]
\hline\\[-6pt]
submit. 4 &  & \checkmark & \checkmark & \checkmark & \textbf{0.471} & 0.480 & 0.477  & 0.479 \\[2pt]\hline\\[-6pt]
submit. 5 &  \checkmark  &  & \checkmark & \checkmark & 0.451 & 0.524 & 0.472 & \textbf{0.498} \\[2pt]
\hline\\[-6pt]
submit. 6 &  \checkmark  & \checkmark & \checkmark & \checkmark & 0.458 & 0.506 & 0.467  & 0.487 \\[2pt]\hline\\[-6pt]
\multicolumn{9}{r}{enet.) using EfficientNet, bal.) data balancing}\\
\multicolumn{9}{r}{ext.) feature extraction, 3s.) using 3s time window}
\end{tabular}
\end{center}
\end{table}

\section{CONCLUSIONS AND FUTURE WORKS}

This paper describes the multi-term and multi-task analyzing framework for estimation of emotion classifications and valence-arousal intensity using the Aff-Wild2 dataset. We introduced multiple time windows into our analyzing framework for extracting feature parameters from video data. Single task models for valence, arousal and expression estimation are generated by ensembling the multiple-term models respectively. Moreover, we fussed the single task models together to furtherly improve the estimation accuracy. The data balancing also was conducted to improve the model robustness. The verification results reveal that our proposed framework has achieved significantly higher performance than baseline on tracks 1 and 2 of the ABAW Challenge.
\\
In the future, we will investigate effective data augmentation including pseudo label and time-series data augmentation, etc. to improve the robustness furtherly.

\end{document}